\pdfoutput=1

\documentclass[11pt]{article}

\usepackage{emnlp2021}
\usepackage{times}
\usepackage{latexsym}
\usepackage{graphicx}

\usepackage[T1]{fontenc}

\usepackage[utf8]{inputenc}

\usepackage{microtype}

\usepackage{inconsolata}

\usepackage{adjustbox}
\usepackage{forest}
\usepackage{microtype}
\usepackage{pifont}
\usepackage{latexsym}

\usepackage{balance}
\usepackage{helvet}  
\usepackage{courier}  
\usepackage{url}  
\usepackage{graphicx}  
\usepackage{amssymb}
\usepackage{amsmath}
\usepackage{algorithm}
\usepackage{algorithmic}
\usepackage{amsthm}
\usepackage{multirow}
\usepackage{pgffor}
\usepackage{graphicx}
\usepackage{epstopdf}
\usepackage{tikz}
\usepackage{epstopdf}
\usepackage{mathtools}
\usepackage{enumitem}
\usepackage{subfigure}
\usepackage{tabulary}
\usepackage{color}
\usepackage{url}
\usepackage{flushend}
\usepackage{stfloats}
\usepackage{tcolorbox}
\usepackage[utf8]{inputenc}
\usepackage[english]{babel}
\usepackage{tabularx}
\usepackage{CJKutf8}
\usepackage{microtype}
\usepackage{textcomp}
\usepackage{booktabs}
\usepackage{colortbl}
\usepackage{soul}
\usepackage{xcolor}
\usepackage{arydshln}
\usepackage{natbib}
\usepackage{appendix}
\usepackage{tabularray}

\title{Investigating Answerability of LLMs for Long-Form Question Answering}

\author{Meghana Moorthy Bhat $\quad$  Rui Meng $\quad$ Ye Liu $\quad$ Yingbo Zhou \and Semih Yavuz\\
Salesforce Research \\
\texttt{\{meghana.bhat, ruimeng, yeliu, yingbo.zhou, syavuz\}@salesforce.com} \\
}
\begin{document}
\maketitle

\begin{abstract}
\vspace{-0.3cm}
As we embark on a new era of LLMs, it becomes increasingly crucial to understand their capabilities, limitations, and differences. 
Toward making further progress in this direction, we strive to build a deeper understanding of the gaps between massive LLMs (e.g., ChatGPT) and smaller yet effective open-source LLMs and their distilled counterparts. 
To this end, we specifically focus on long-form question answering (LFQA) because it has several practical and impactful applications (e.g., troubleshooting, customer service, etc.) yet is still understudied and challenging for LLMs. 
We propose a question-generation method from abstractive summaries and show that generating follow-up questions from summaries of long documents can create a challenging setting for LLMs to reason and infer from long contexts. 
Our experimental results confirm that: (1) our proposed method of generating questions from abstractive summaries pose a challenging setup for LLMs and shows performance gaps between LLMs like ChatGPT and open-source LLMs (Alpaca, Llama) 
(2) open-source LLMs exhibit decreased reliance on context for generated questions from the original document, but their generation capabilities drop significantly on generated questions from summaries -- especially for longer contexts (>1024 tokens).

\end{abstract}

\vspace{-0.5cm}
\section{Introduction}
\vspace{-0.2cm}
While Large Language Models (LLMs) like ChatGPT, GPT-4 \cite{openai2023gpt4} have exhibited superior performance across various benchmarks, open-source efforts have also been progressing rapidly in catching up across different applications and benchmarks like MMLU \cite{hendryckstest2021}, OpenLLMBoard \cite{anil2023palm,open-llm-leaderboard,touvron2023llama}. 
As we move into the new era of LLMs with fast-paced progress on new models and techniques, it becomes increasingly important to understand the capabilities, limitations, and differences between them.

With LLMs capable of generating coherent text has proven to perform well in tasks like summarization \cite{ouyang2022training}, their performance on LFQA is relatively less known. Long-Form Question Answering (LFQA) is one of the important unsolved challenges with diverse and impactful real-world applications (e.g., help forums, troubleshooting, customer services, etc.)
Answering such questions often requires complex reasoning abilities to understand query and reason across spans of information scattered across original document. 


Abstractive summaries contain the crux of the articles in a compressed form \cite{fabbri2020summeval}. We hypothesize that follow-up questions from these summaries would require a deeper understanding of the topics that would link different parts of the source document. 
Moreover, \citet{pang-etal-2022-quality} demonstrate that answers that require understanding more than a third of the long document are often rated as ``HARD'' by humans. 
Therefore, we propose a scalable evaluation method to analyze and study the disparities of massive LLMs with smaller yet proven successful base LLMs (e.g., Llama-7B, 13B) and their distilled versions (e.g., Alpaca-7B, 13B). 
To this end, we propose to prompt ChatGPT with specific instructions to generate complex questions from document summaries. 

Our empirical analysis on two fronts (complexity of generated questions and answer quality of open-source LLMs) show that follow-up questions generated from summaries pose a challenging yet more realistic setup for testing the reasoning abilities of LLMs. 
Since relying fully on the human evaluation for long-form QA is expensive and difficult to scale \cite{pagnoni-etal-2021-understanding}, we instead leverage GPT-4 to evaluate the answer quality on coherence, relevance, factual consistency, and accuracy following prior works \cite{fabbri2020summeval,fan-etal-2019-eli5}.
However, we also do a smaller scale human evaluation, which shows that GPT-4 has a high correlation with human evaluation making our evaluation reliable. 
Our main findings from this study are as follows:
\vspace{-0.2cm}
\begin{itemize}[leftmargin=*]
\vspace{-0.35cm}
    \item Our proposed method of generating questions from abstractive summaries require inferring from longer contexts, with multiple passes through the context for > 20\% times. 
    \vspace{-0.5cm}
    \item Distilled LLMs (Alpaca-7B, 13B) tend to rely less on context for generated questions from the original document, but their generation capabilities drop significantly on generated questions from document summaries. 
    \vspace{-0.5cm}
    \item Answers generated by distilled LLMs can be coherent across different settings; but tend to drift from the question, generate repetitive and partially correct answers for the questions generated from summaries (> 16.8\%) 
    \vspace{-0.5cm}
    \item Alpaca-7B, 13B mostly generate meaningful answers than base LLMs (Llama) but are sensitive to longer contexts (>1024 tokens).
    \vspace{-0.1cm}
\end{itemize}

\section{Related Work}
\textbf{Reasoning over Long Documents:} LLMs have shown amazing capabilities to reason over a number of tasks like commonsense reasoning \cite{talmor-etal-2019-commonsenseqa}, mathematical and symbolic reasoning \cite{huang2023reasoning,cobbe2021training}, question answering tasks like SQuaD, HotpotQA. However, most of these tasks do not require long context and answers are often a short phrase or a span of text from the context. 
In this work, we evaluate LLMs to reason over long documents that would require deeper understanding capabilities and longer context to answer by prompting LLMs (ChatGPT) to generate follow-up questions from summaries of long documents. \\
\textbf{Model-based Evaluation}: Prior work has proposed automatic evaluation metrics using learned models \cite{Zhang*2020BERTScore:,laban-etal-2022-summac}; especially for long form text generation tasks like summarization \cite{fabbri2020summeval,kryscinski-etal-2020-evaluating} where consistency and coherency is measured between the source document and generated summary as entailment. Recently, \citet{liu2023geval} showed that GPT-4 has the highest correlation with humans and surpasses all other auto-evaluation methods on summarization tasks. We take inspiration from these works to design evaluation prompts and use GPT-4 as the evaluator for generated answers. 
\section{Proposed Evaluation Method} 
\subsection{Data Source}
\label{sec:data_src}

In order to create a dataset that is diverse and widely usable, we use Wikipedia articles. Using metadata of the Category list from Wikipedia API, we control the diversity of topics and collect articles from each of the following 9 domains: Physics, Entertainment, History, Computer Science, Social Sciences, Society, Economics, Medicine, and Sports. \\
The article pages can often be lengthy to fit in the context of LLMs. Hence, we extract section-wise data from the article pages that have a minimum length of 256 tokens using SpaCy tokenizer and combine the shorter sections together. For a fair comparison between different models, we set a maximum context length of 2k tokens in our experiments. In addition, we filter out non-informative documents using pre-processing filters. Further details are available in Appendix ~\ref{app:data_source}.
\subsection{Question Generation using ChatGPT}
\vspace{-0.1cm}
We formulate our question generation method as a two-step process: (1) Summarization and (2) Question generation from summary. \\
\textbf{Summarization } First, we collect section wise passages from Wikipedia as described in Section ~\ref{sec:data_src}. Then, we prompt ChatGPT (\textbf{gpt-turbo-3.5}) to generate summary of original document. In order to provide more context to ChatGPT, we provide information about the title and the domain of the article in the passage. \\
\textbf{Question generation from summary } In this step, we prompt ChatGPT to generate questions using document summaries as context. To avoid random order question generation, we instruct ChatGPT to provide top-3 complex questions to answer.
To demonstrate the usefulness of our question generation process, we also establish a baseline with the same instructions where questions are directly generated from the passage. \\
\indent Please refer to the appendix ~\ref{app:question_generation} for the prompt used in our setup. In summary, we generate 3 questions for 50 passages in each domain totaling to \texttt{1350 questions} for each setting. 
\subsection{Evaluation of Generated Question Complexity}
\begin{table}[h!]
\footnotesize
\centering
\resizebox{0.5\textwidth}{!}{%
\begin{tabular}{ll}
\toprule
\textbf{Question} & \textbf{Choices} \\
\hline
Q1: Is the question answerable from the  & \\
given context and is unambiguous? & A. Yes B. No\\
\midrule
Q2. How much of the passage is needed  & A. Only a sentence or two \\
as context to answer the question? & B. More than 2 sentences \\
& but lesser than a paragraph  \\
& C. Atleast a third of the \\
& entire passage \\
& D. Most of the passage \\ 
\midrule
Q3: Does the question require multiple & \\
passes through the passage? & A. Yes B. No \\
\bottomrule
\end{tabular}
}
\caption{Prompts designed to evaluate the complexity of generated questions.} 
\label{tab:prompts_QG_eval_questions}
\vspace{-0.75cm}
\end{table}

\citet{pang-etal-2022-quality} designed extensive annotation guidelines to assess the complexity of questions. Of the questions rated as 'HARD' by humans, 26.7\% of the questions (20.2\% higher than the easier ones) needed at least one-third or more of the given information to be answered.  In order to assess the quality of generated questions, we prompt ChatGPT with the questions (Table ~\ref{tab:prompts_QG_eval_questions}) for (1) From the passage (\texttt{QG-Passage}) (2) From the summary (\texttt{QG-Summary}). Following prior work, by majority voting we exclude the questions that are rated as unanswerable by ChatGPT by prompting the questions with different \texttt{top\_p = \{0.8, 0.9, 1\}}. After filtering, we have \texttt{1278} generated questions from each setting.
\vspace{-0.75cm}
\begin{table}[h!]
\footnotesize
\resizebox{0.5\textwidth}{!}{%
\begin{tabular}{lll}
\toprule
\textbf{Evaluation Metric} & \textbf{QG - Passage} & \textbf{QG - Summary}\\
\hline
Q1: Unambiguity  & 96.6\% & 94.7\% \\
\midrule
Q2. Context Length:  &  & \\
\quad A sentence or less than a paragraph  & \textbf{79.3\%} & 75.7 \% \\
\quad At least a third or most of the passage & 20.7\% & \textbf{24.3\%} \\ 
\midrule
Q3: Multi-pass of the passage & 24.4\% & 31\% \\
\bottomrule
\end{tabular}
}
\caption{Prompts designed to evaluate the complexity of generated questions.} 
\label{tab:prompts_QG_eval}
\end{table}


\section{Results and Analysis}
\label{sec:results}
\subsection{Experiment Setup}
As few-shot setting is infeasible in our setting due to context length, we compare model performance on zero-shot evaluation. We prompt the following models to generate free-form text as answers on our final evaluation dataset: ChatGPT \cite{openai2023gpt4}, Alpaca-7B, 13B \cite{alpaca}, LLaMa-7B, 13B \cite{touvron2023llama}. We use OpenAI API for ChatGPT and load checkpoints for open-source LLMs from HuggingFace \footnote{https://huggingface.com}. The prompt used for generating answers are in Appendix ~\ref{appendix}. Please note that our experiments do not consider input beyond 2k sequence length for fair comparisons with other models. We also test generating questions from Alpaca and found them to not follow instructions and often generate irrelevant content. Our detailed analysis can be found in Appendix~\ref{app:alpaca_QG}. \\
\indent \textbf{GPT-4 as evaluator} has shown high correlation with human evaluation in long form text generation tasks like summarization \cite{liu2023geval} surpassing other auto-evaluation metrics like ROUGE and BLEU scores. 
Since LLMs are expected to generate free form answers for our setting, we take inspiration from prior works on long-form text generation metrics \cite{fabbri2020summeval} and adopt them in our evaluation for coherency, consistency, accuracy, and relevance. Basically, we adopt the definitions used as guidelines for human evaluation to our method as shown below: \\
\textbf{Coherency:} Answer should be well-structured
and well-organized and should not just
be a heap of related information. \\
\textbf{Relevance:} Answer should be relevant to the question and the context. The answer should be concise and avoid drifting from the question being asked. \\
\textbf{Factual consistency:} The context should be the primary source for the answer. The answer should not contain fabricated facts and should entail information present in the context. \\
\textbf{Accuracy:} Answer should be satisfactory and complete to the question being asked. Measure the correctness of the answer by checking if the response answers the presented question. \\
\indent We prompt GPT-4 to rate answers on a scale from 0 to 3 (higher the better) on all of the four metrics. We average all the ratings obtained from GPT-4 and present the results in Table ~\ref{tab: main results}. Our evaluation prompt can be found in Appendix~\ref{app:answer_evaluation}. \\
\indent We hypothesize that an optimal prompt should always prefer human answers and not be biased towards model-generated answers. \citet{laskar2023systematic} show that LLMs like ChatGPT still underperform to humans on TruthfulQA dataset\cite{lin2022truthfulqa}. Hence, we perform proxy testing with GPT-4 on TruthfulQA dataset in order to verify the reliability and faithfulness of our evaluation prompt. We test the generated answers from ChatGPT and open-source LLMs against the ground truth on randomly sampled 50 test instances and find that our evaluation prompt with GPT-4 prompt prefers human-written answers for factual consistency and correctness over model-generated ones more than > 90\% of the times. 
In addition, we also perform human evaluation of LLM generated answers and discuss the correlation of GPT-4 evaluation with human evaluation in Section~\ref{app:human_eval}. 
\vspace{-0.3cm}
\subsection{Results}
\begin{table}[ht]
\centering
\resizebox{0.5\textwidth}{!}{%
\begin{tabular}{lccccc}
\toprule
\textbf{Model}
& \multicolumn{2}{c}{\textbf{QG-Passage}}
& \multicolumn{2}{c}{\textbf{QG-Summary}} \\
& \small{w/o context} & \small{w/ context}
& \small{w/o context} & \small{w/ context}
& \quad \\
\midrule
\,ChatGPT & \textbf{2.78} & \textbf{2.93} & \textbf{2.67} & \textbf{2.82}   \\ 
\midrule
\,Alpaca-13B  & 2.27 & 2.09 & 2.04 & 2.09    \\ 
\,LlaMa-13B & 1.22 & 1.47 & 0.98 &  1.28  \\ 
\midrule
\,Alpaca-7B & 2.04 & 1.96 & 1.64 &  1.89   \\ 
\,LlaMa-7B & 0.89 & 1.12 & 0.66 &  0.78   \\ 
\bottomrule
\end{tabular}
}
 \vspace{-0.5em} \caption{ Performance of different models based on GPT-4 evaluation. The table shows average ratings across all metrics: accuracy, coherency, consistency, relevance.
}
 \vspace{-1em}
\label{tab: main results}
\vspace{-0.5cm}
\end{table}

\begin{figure*}
    \centering
    \subfigure{\includegraphics[width=0.24\textwidth]{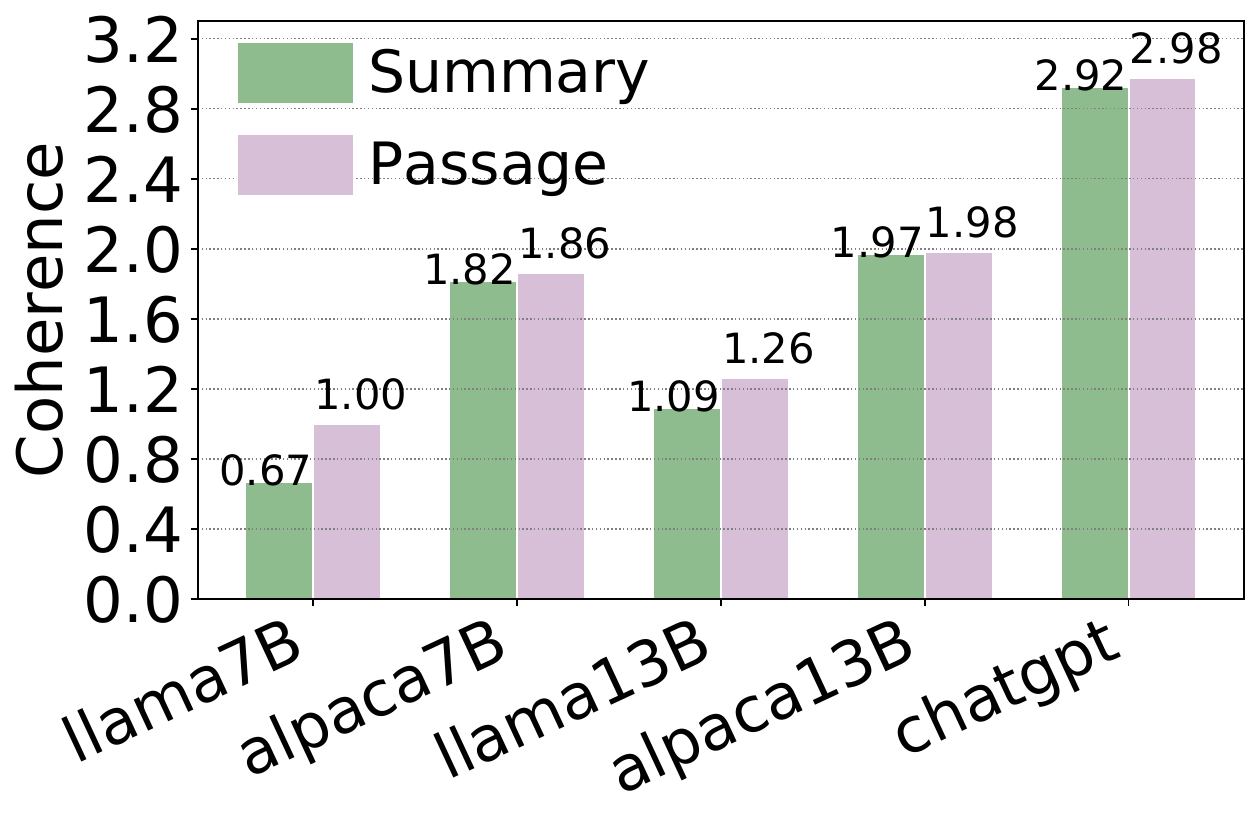}} 
    \subfigure{\includegraphics[width=0.24\textwidth]{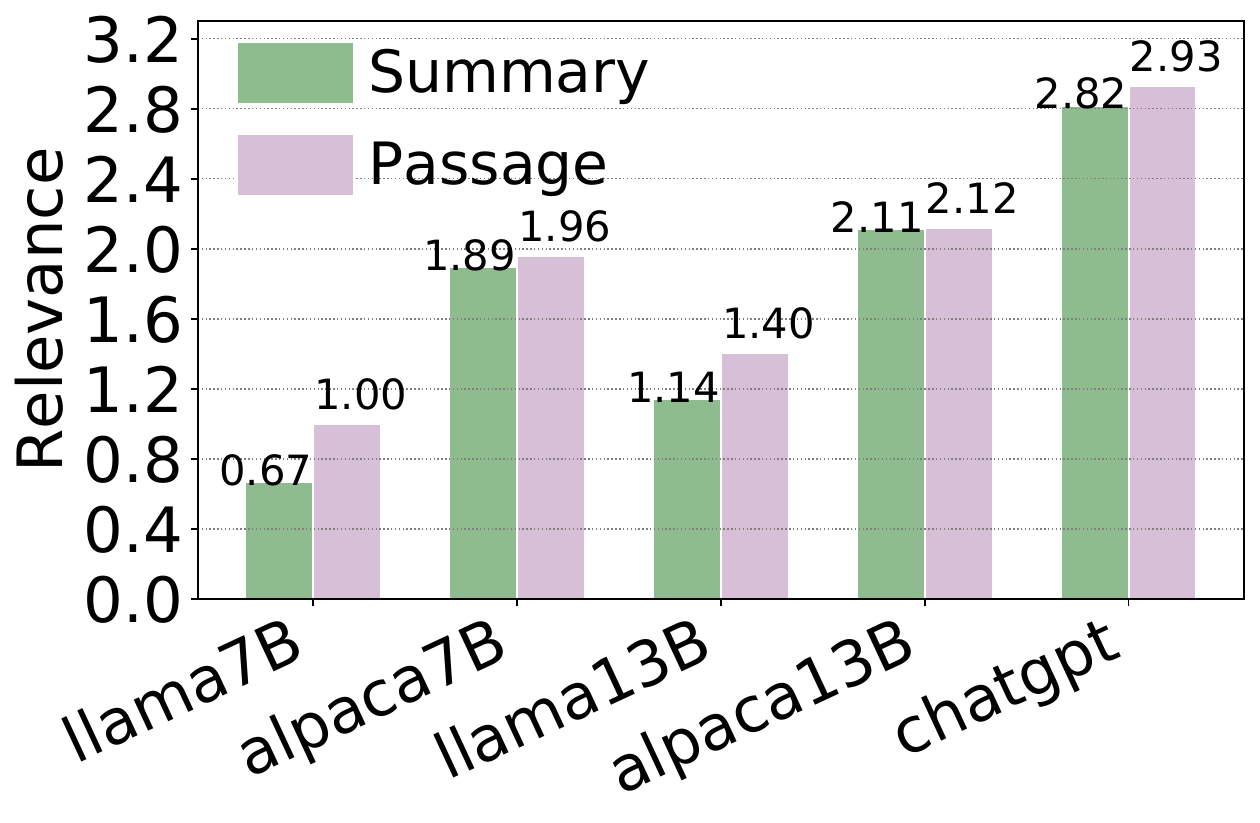}}
    \subfigure{\includegraphics[width=0.24\textwidth]{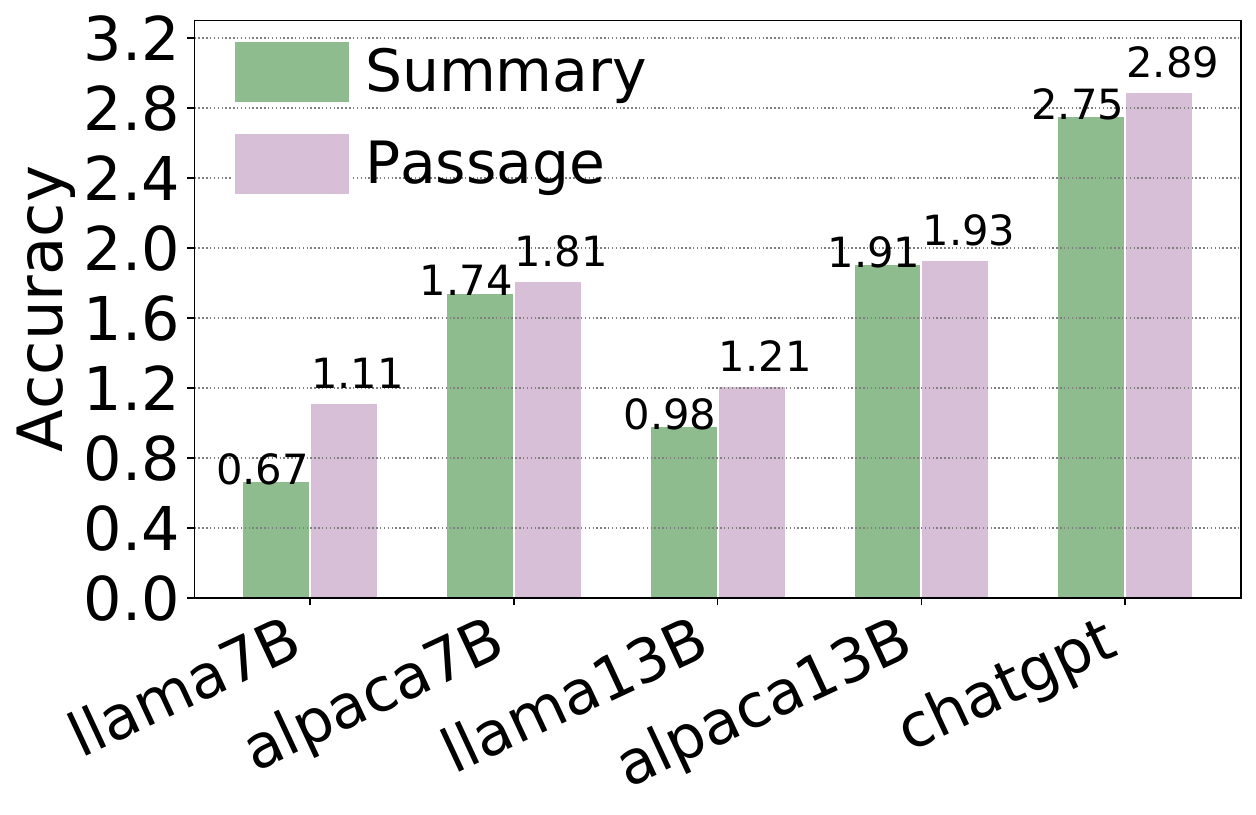}}
    \subfigure{\includegraphics[width=0.24\textwidth]{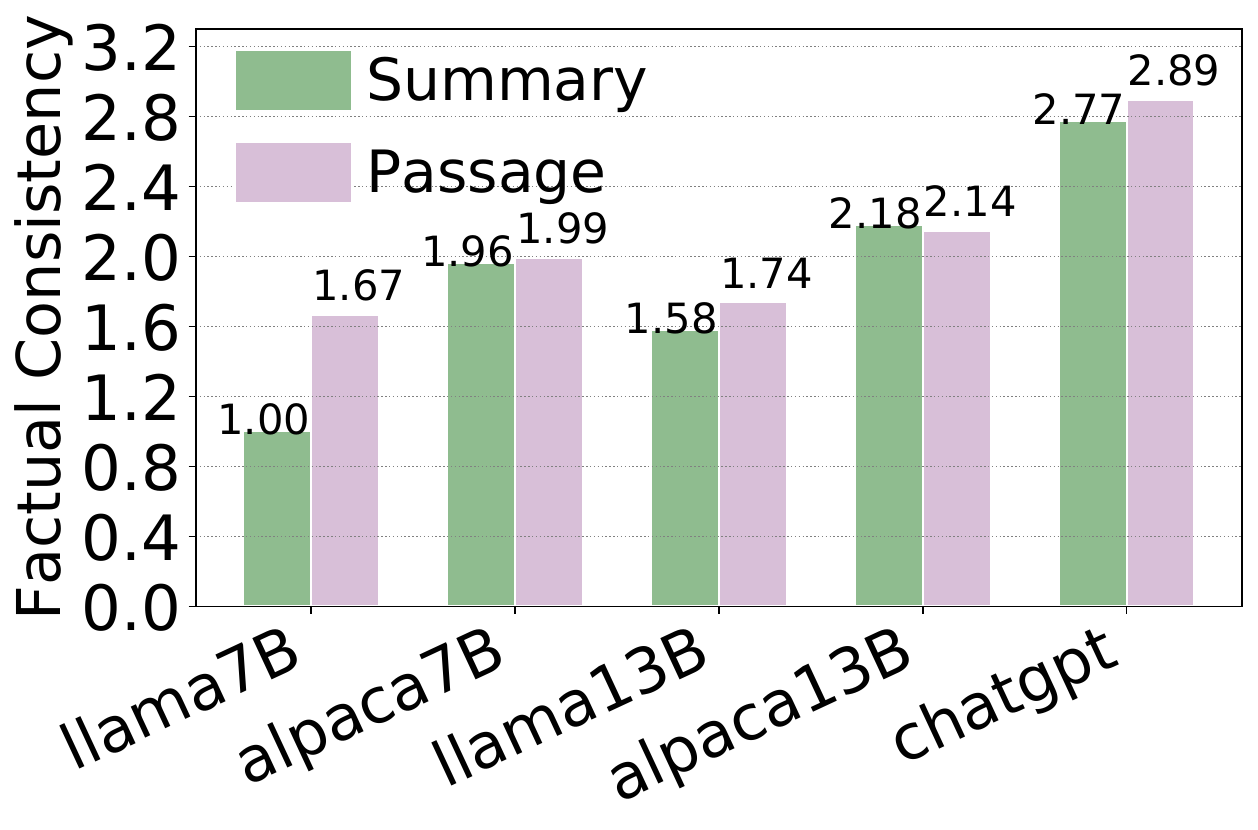}}
    \vspace{-0.55cm} \caption{Graphs showing the breakdown of models with respect to different metrics used in evaluation: (a) Coherence (b) Relevance (c) Answer Accuracy (d) Factual Consistency}
    \label{fig:metrics_breakdown}
    \vspace{-0.5cm}
\end{figure*}

\begin{figure}
    \centering
    \includegraphics[width=0.4\textwidth]{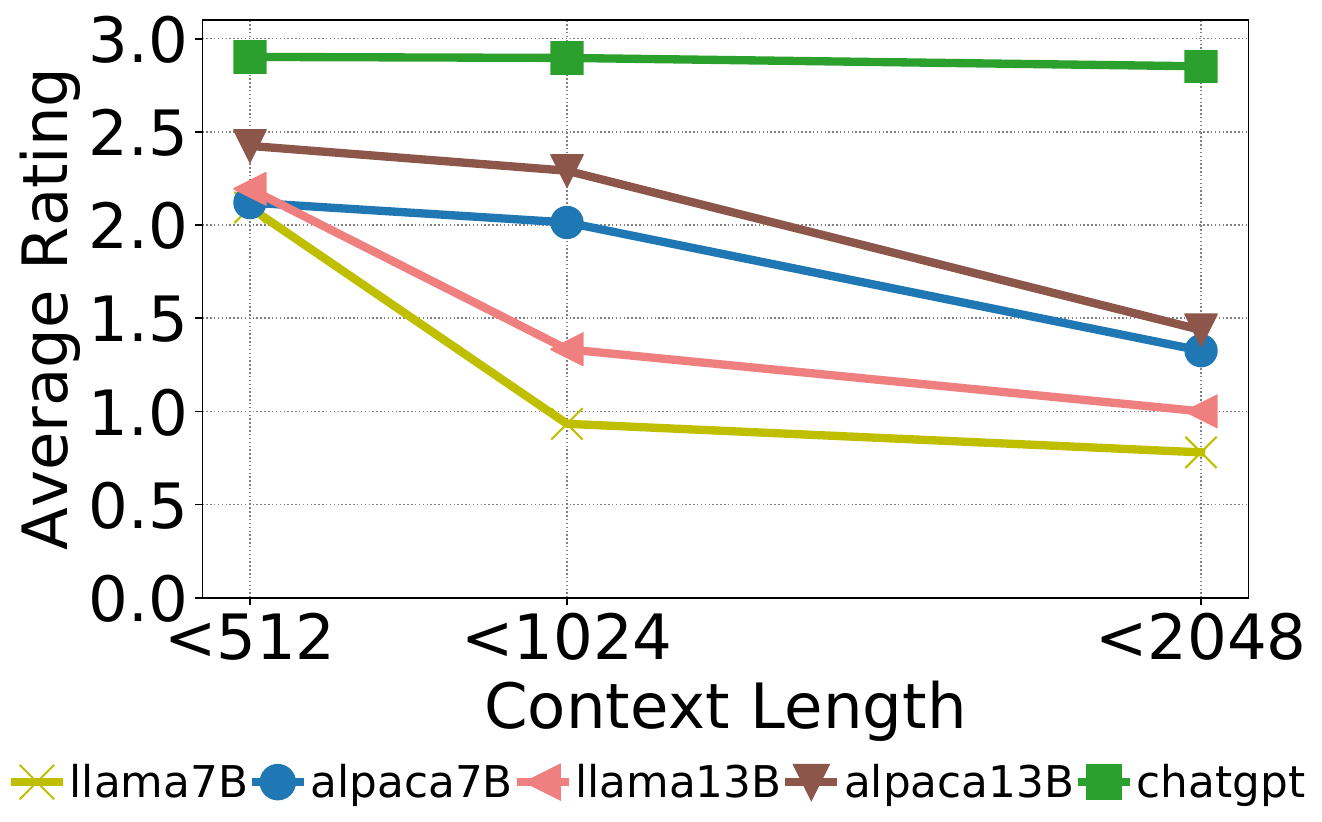}
    \vspace{-0.3cm}
    \caption{Performance (avg. ratings) of LLMs across different context length.}
    \label{fig:context_length_analysis}
    \vspace{-0.75cm}
\end{figure}

Our experiment results show that ChatGPT outperforms other LLMs in all the metrics by a wide margin from \textit{22.4\% - 40.1\%} against the second-best performing LLM (Alpaca-13B). However; all the models including ChatGPT generate less accurate and relevant answers for \texttt{QG-Summary} when compared to \texttt{QG-Passage}; while the gap is much larger in open-source LLMs.
We also find that most of the LLMs find context important in order to generate answers; however, the gap is much smaller for \texttt{QG-Passage} (avg. gap of 0.12 v.s. 0.2). Surprisingly, Alpaca-7B, 13B models perform better w/o context for \texttt{QG-Passage}. We hypothesize that questions directly generated from the context passage can be simple that could be directly answered from the parametric knowledge of LLMs without additional context. On further analysis, we observe that Alpaca-7B,13B performance drops significantly in longer contexts (Figure \ref{fig:context_length_analysis}). We hypothesize that in a constrained sequence length setting, adding supporting context (even gold passage) may not be always helpful. We leave further analysis for future work and hope our analysis will motivate future research directions to study when to add/not add context for different tasks. \\
\textbf{Performance of LLMs on different metrics } Figure~\ref{fig:metrics_breakdown} presents the performance of models across different metrics for QG-Summary. We observe two trends: (1) Open-source base LLMs (Llama-7B,13B) suffer at all fronts significantly on generated answer quality whereas distilled models perform better than their counterparts (Llama) on all the settings. (2) \texttt{QG-Summary} provides a more challenging setting for all the LLMs: specifically, we notice that degradation in coherency score is negligent on ChatGPT and Alpaca-13B while other metrics like relevance, answer accuracy and factual consistency degrade consistently. We find open-source LLMs to drift from the question, generate partially correct answers and repeat more frequently in QG-Summary setting leading to lower scores. This further confirms that our proposed evaluation method \texttt{QG-Summary} challenges LLMs for deeper reasoning capabilities. //
\textbf{Context Length Analysis } We analyze the effect of context length across LLMs in our proposed setting (\texttt{QG-Summary}). As expected, ChatGPT remains robust to context length until 2k tokens with Llama variants performing worse than other models (Figure \ref{fig:context_length_analysis}). Interestingly, we find distilled models (Alpaca) being consistent until 1024 tokens, however beyond > 1024 tokens, the performance degrades at a higher rate than Llama.

\vspace{-0.3cm}
\section{Conclusion}
\vspace{-0.3cm}
With the emergence of LLMs like ChatGPT and open-source successful LLMs, it is extremely important to understand the capabilities and limitations of different LLMs. In order to test deeper reasoning abilities of LLMs by referring to longer contexts, we evaluate answers generated by LLMs on questions generated by ChatGPT on summaries of long documents. Results show that our proposed method of question generation poses a challenging setup for LLMs and shed light on performance gaps between massive LLMs and open-source LLMs. 
We hope our analysis motivates future research directions such as leveraging longer contexts in a constrained sequence length setting and developing better long-form text generation for smaller LLMs.
\section{Limitations}
In this study, we propose an automatic evaluation setting to generate questions from summaries, and the generated answers from LLMs are evaluated using GPT-4 for different metrics. Experimental results show that our proposed evaluation setting proves to be a challenging setup for LLMs. However, our study might have some limitations. \\
\textbf{GPT-4 as evaluator } While GPT-4 has shown a high correlation with human evaluation for long form text generation \cite{liu2023geval}, the capabilities of using GPT-4 for evaluation is an active area of research in itself. Hence, our results might be limited by the undiscovered capabilities of GPT-4. \\
\textbf{ChatGPT for question generation } Generating answers on questions prompted from ChatGPT might lead to optimistic results of ChatGPT. However, there exists limitations with other baselines to generate meaningful questions. We show extensive analysis of using other LLMs for question generation (Appendix~\ref{app:alpaca_QG}). \\
\textbf{Unknown training data } Little is known about the training data distribution of massive LLMs like ChatGPT. Models trained with different methods and data distribution make the evaluation for fair comparison harder. 
\bibliography{anthology}

\begin{thebibliography}{21}
\expandafter\ifx\csname natexlab\endcsname\relax\def\natexlab#1{#1}\fi

\bibitem[{Anil et~al.(2023)Anil, Dai, Firat, Johnson, Lepikhin, Passos,
  Shakeri, Taropa, Bailey, Chen, Chu, Clark, Shafey, Huang, Meier-Hellstern,
  Mishra, Moreira, Omernick, Robinson, Ruder, Tay, Xiao, Xu, Zhang, Abrego,
  Ahn, Austin, Barham, Botha, Bradbury, Brahma, Brooks, Catasta, Cheng, Cherry,
  Choquette-Choo, Chowdhery, Crepy, Dave, Dehghani, Dev, Devlin, Díaz, Du,
  Dyer, Feinberg, Feng, Fienber, Freitag, Garcia, Gehrmann, Gonzalez, Gur-Ari,
  Hand, Hashemi, Hou, Howland, Hu, Hui, Hurwitz, Isard, Ittycheriah, Jagielski,
  Jia, Kenealy, Krikun, Kudugunta, Lan, Lee, Lee, Li, Li, Li, Li, Li, Lim, Lin,
  Liu, Liu, Maggioni, Mahendru, Maynez, Misra, Moussalem, Nado, Nham, Ni,
  Nystrom, Parrish, Pellat, Polacek, Polozov, Pope, Qiao, Reif, Richter, Riley,
  Ros, Roy, Saeta, Samuel, Shelby, Slone, Smilkov, So, Sohn, Tokumine, Valter,
  Vasudevan, Vodrahalli, Wang, Wang, Wang, Wang, Wieting, Wu, Xu, Xu, Xue, Yin,
  Yu, Zhang, Zheng, Zheng, Zhou, Zhou, Petrov, and Wu}]{anil2023palm}
Rohan Anil, Andrew~M. Dai, Orhan Firat, Melvin Johnson, Dmitry Lepikhin,
  Alexandre Passos, Siamak Shakeri, Emanuel Taropa, Paige Bailey, Zhifeng Chen,
  Eric Chu, Jonathan~H. Clark, Laurent~El Shafey, Yanping Huang, Kathy
  Meier-Hellstern, Gaurav Mishra, Erica Moreira, Mark Omernick, Kevin Robinson,
  Sebastian Ruder, Yi~Tay, Kefan Xiao, Yuanzhong Xu, Yujing Zhang,
  Gustavo~Hernandez Abrego, Junwhan Ahn, Jacob Austin, Paul Barham, Jan Botha,
  James Bradbury, Siddhartha Brahma, Kevin Brooks, Michele Catasta, Yong Cheng,
  Colin Cherry, Christopher~A. Choquette-Choo, Aakanksha Chowdhery, Clément
  Crepy, Shachi Dave, Mostafa Dehghani, Sunipa Dev, Jacob Devlin, Mark Díaz,
  Nan Du, Ethan Dyer, Vlad Feinberg, Fangxiaoyu Feng, Vlad Fienber, Markus
  Freitag, Xavier Garcia, Sebastian Gehrmann, Lucas Gonzalez, Guy Gur-Ari,
  Steven Hand, Hadi Hashemi, Le~Hou, Joshua Howland, Andrea Hu, Jeffrey Hui,
  Jeremy Hurwitz, Michael Isard, Abe Ittycheriah, Matthew Jagielski, Wenhao
  Jia, Kathleen Kenealy, Maxim Krikun, Sneha Kudugunta, Chang Lan, Katherine
  Lee, Benjamin Lee, Eric Li, Music Li, Wei Li, YaGuang Li, Jian Li, Hyeontaek
  Lim, Hanzhao Lin, Zhongtao Liu, Frederick Liu, Marcello Maggioni, Aroma
  Mahendru, Joshua Maynez, Vedant Misra, Maysam Moussalem, Zachary Nado, John
  Nham, Eric Ni, Andrew Nystrom, Alicia Parrish, Marie Pellat, Martin Polacek,
  Alex Polozov, Reiner Pope, Siyuan Qiao, Emily Reif, Bryan Richter, Parker
  Riley, Alex~Castro Ros, Aurko Roy, Brennan Saeta, Rajkumar Samuel, Renee
  Shelby, Ambrose Slone, Daniel Smilkov, David~R. So, Daniel Sohn, Simon
  Tokumine, Dasha Valter, Vijay Vasudevan, Kiran Vodrahalli, Xuezhi Wang,
  Pidong Wang, Zirui Wang, Tao Wang, John Wieting, Yuhuai Wu, Kelvin Xu, Yunhan
  Xu, Linting Xue, Pengcheng Yin, Jiahui Yu, Qiao Zhang, Steven Zheng,
  Ce~Zheng, Weikang Zhou, Denny Zhou, Slav Petrov, and Yonghui Wu. 2023.
\newblock \href {http://arxiv.org/abs/2305.10403} {Palm 2 technical report}.

\bibitem[{Beeching et~al.(2023)Beeching, Han, Lambert, Rajani, Sanseviero,
  Tunstall, and Wolf}]{open-llm-leaderboard}
Edward Beeching, Sheon Han, Nathan Lambert, Nazneen Rajani, Omar Sanseviero,
  Lewis Tunstall, and Thomas Wolf. 2023.
\newblock Open llm leaderboard.
\newblock
  \url{https://huggingface.co/spaces/HuggingFaceH4/open_llm_leaderboard}.

\bibitem[{Cobbe et~al.(2021)Cobbe, Kosaraju, Bavarian, Chen, Jun, Kaiser,
  Plappert, Tworek, Hilton, Nakano, Hesse, and Schulman}]{cobbe2021training}
Karl Cobbe, Vineet Kosaraju, Mohammad Bavarian, Mark Chen, Heewoo Jun, Lukasz
  Kaiser, Matthias Plappert, Jerry Tworek, Jacob Hilton, Reiichiro Nakano,
  Christopher Hesse, and John Schulman. 2021.
\newblock \href {http://arxiv.org/abs/2110.14168} {Training verifiers to solve
  math word problems}.

\bibitem[{Fabbri et~al.(2020)Fabbri, Kry{\'s}ci{\'n}ski, McCann, Xiong, Socher,
  and Radev}]{fabbri2020summeval}
Alexander~R Fabbri, Wojciech Kry{\'s}ci{\'n}ski, Bryan McCann, Caiming Xiong,
  Richard Socher, and Dragomir Radev. 2020.
\newblock Summeval: Re-evaluating summarization evaluation.
\newblock \emph{arXiv preprint arXiv:2007.12626}.

\bibitem[{Fan et~al.(2019)Fan, Jernite, Perez, Grangier, Weston, and
  Auli}]{fan-etal-2019-eli5}
Angela Fan, Yacine Jernite, Ethan Perez, David Grangier, Jason Weston, and
  Michael Auli. 2019.
\newblock \href {https://doi.org/10.18653/v1/P19-1346} {{ELI}5: Long form
  question answering}.
\newblock In \emph{Proceedings of the 57th Annual Meeting of the Association
  for Computational Linguistics}, pages 3558--3567, Florence, Italy.
  Association for Computational Linguistics.

\bibitem[{Hendrycks et~al.(2021)Hendrycks, Burns, Basart, Zou, Mazeika, Song,
  and Steinhardt}]{hendryckstest2021}
Dan Hendrycks, Collin Burns, Steven Basart, Andy Zou, Mantas Mazeika, Dawn
  Song, and Jacob Steinhardt. 2021.
\newblock Measuring massive multitask language understanding.
\newblock \emph{Proceedings of the International Conference on Learning
  Representations (ICLR)}.

\bibitem[{Huang and Chang(2023)}]{huang2023reasoning}
Jie Huang and Kevin Chen-Chuan Chang. 2023.
\newblock \href {http://arxiv.org/abs/2212.10403} {Towards reasoning in large
  language models: A survey}.

\bibitem[{Ko{\v{c}}isk{\'y} et~al.(2018)Ko{\v{c}}isk{\'y}, Schwarz, Blunsom,
  Dyer, Hermann, Melis, and Grefenstette}]{kocisky-etal-2018-narrativeqa}
Tom{\'a}{\v{s}} Ko{\v{c}}isk{\'y}, Jonathan Schwarz, Phil Blunsom, Chris Dyer,
  Karl~Moritz Hermann, G{\'a}bor Melis, and Edward Grefenstette. 2018.
\newblock \href {https://doi.org/10.1162/tacl_a_00023} {The {N}arrative{QA}
  reading comprehension challenge}.
\newblock \emph{Transactions of the Association for Computational Linguistics},
  6:317--328.

\bibitem[{Kryscinski et~al.(2020)Kryscinski, McCann, Xiong, and
  Socher}]{kryscinski-etal-2020-evaluating}
Wojciech Kryscinski, Bryan McCann, Caiming Xiong, and Richard Socher. 2020.
\newblock \href {https://doi.org/10.18653/v1/2020.emnlp-main.750} {Evaluating
  the factual consistency of abstractive text summarization}.
\newblock In \emph{Proceedings of the 2020 Conference on Empirical Methods in
  Natural Language Processing (EMNLP)}, pages 9332--9346, Online. Association
  for Computational Linguistics.

\bibitem[{Laban et~al.(2022)Laban, Schnabel, Bennett, and
  Hearst}]{laban-etal-2022-summac}
Philippe Laban, Tobias Schnabel, Paul~N. Bennett, and Marti~A. Hearst. 2022.
\newblock \href {https://doi.org/10.1162/tacl_a_00453} {{S}umma{C}: Re-visiting
  {NLI}-based models for inconsistency detection in summarization}.
\newblock \emph{Transactions of the Association for Computational Linguistics},
  10:163--177.

\bibitem[{Laskar et~al.(2023)Laskar, Bari, Rahman, Bhuiyan, Joty, and
  Huang}]{laskar2023systematic}
Md~Tahmid~Rahman Laskar, M~Saiful Bari, Mizanur Rahman, Md~Amran~Hossen
  Bhuiyan, Shafiq Joty, and Jimmy~Xiangji Huang. 2023.
\newblock \href {http://arxiv.org/abs/2305.18486} {A systematic study and
  comprehensive evaluation of chatgpt on benchmark datasets}.

\bibitem[{Lin et~al.(2022)Lin, Hilton, and Evans}]{lin2022truthfulqa}
Stephanie Lin, Jacob Hilton, and Owain Evans. 2022.
\newblock \href {http://arxiv.org/abs/2109.07958} {Truthfulqa: Measuring how
  models mimic human falsehoods}.

\bibitem[{Liu et~al.(2023)Liu, Iter, Xu, Wang, Xu, and Zhu}]{liu2023geval}
Yang Liu, Dan Iter, Yichong Xu, Shuohang Wang, Ruochen Xu, and Chenguang Zhu.
  2023.
\newblock \href {http://arxiv.org/abs/2303.16634} {G-eval: Nlg evaluation using
  gpt-4 with better human alignment}.

\bibitem[{OpenAI(2023)}]{openai2023gpt4}
OpenAI. 2023.
\newblock \href {http://arxiv.org/abs/2303.08774} {Gpt-4 technical report}.

\bibitem[{Ouyang et~al.(2022)Ouyang, Wu, Jiang, Almeida, Wainwright, Mishkin,
  Zhang, Agarwal, Slama, Ray, Schulman, Hilton, Kelton, Miller, Simens, Askell,
  Welinder, Christiano, Leike, and Lowe}]{ouyang2022training}
Long Ouyang, Jeff Wu, Xu~Jiang, Diogo Almeida, Carroll~L. Wainwright, Pamela
  Mishkin, Chong Zhang, Sandhini Agarwal, Katarina Slama, Alex Ray, John
  Schulman, Jacob Hilton, Fraser Kelton, Luke Miller, Maddie Simens, Amanda
  Askell, Peter Welinder, Paul Christiano, Jan Leike, and Ryan Lowe. 2022.
\newblock \href {http://arxiv.org/abs/2203.02155} {Training language models to
  follow instructions with human feedback}.

\bibitem[{Pagnoni et~al.(2021)Pagnoni, Balachandran, and
  Tsvetkov}]{pagnoni-etal-2021-understanding}
Artidoro Pagnoni, Vidhisha Balachandran, and Yulia Tsvetkov. 2021.
\newblock \href {https://doi.org/10.18653/v1/2021.naacl-main.383}
  {Understanding factuality in abstractive summarization with {FRANK}: A
  benchmark for factuality metrics}.
\newblock In \emph{Proceedings of the 2021 Conference of the North American
  Chapter of the Association for Computational Linguistics: Human Language
  Technologies}, pages 4812--4829, Online. Association for Computational
  Linguistics.

\bibitem[{Pang et~al.(2022)Pang, Parrish, Joshi, Nangia, Phang, Chen,
  Padmakumar, Ma, Thompson, He, and Bowman}]{pang-etal-2022-quality}
Richard~Yuanzhe Pang, Alicia Parrish, Nitish Joshi, Nikita Nangia, Jason Phang,
  Angelica Chen, Vishakh Padmakumar, Johnny Ma, Jana Thompson, He~He, and
  Samuel Bowman. 2022.
\newblock \href {https://doi.org/10.18653/v1/2022.naacl-main.391} {{Q}u{ALITY}:
  Question answering with long input texts, yes!}
\newblock In \emph{Proceedings of the 2022 Conference of the North American
  Chapter of the Association for Computational Linguistics: Human Language
  Technologies}, pages 5336--5358, Seattle, United States. Association for
  Computational Linguistics.

\bibitem[{Talmor et~al.(2019)Talmor, Herzig, Lourie, and
  Berant}]{talmor-etal-2019-commonsenseqa}
Alon Talmor, Jonathan Herzig, Nicholas Lourie, and Jonathan Berant. 2019.
\newblock \href {https://doi.org/10.18653/v1/N19-1421} {{C}ommonsense{QA}: A
  question answering challenge targeting commonsense knowledge}.
\newblock In \emph{Proceedings of the 2019 Conference of the North {A}merican
  Chapter of the Association for Computational Linguistics: Human Language
  Technologies, Volume 1 (Long and Short Papers)}, pages 4149--4158,
  Minneapolis, Minnesota. Association for Computational Linguistics.

\bibitem[{Taori et~al.(2023)Taori, Gulrajani, Zhang, Dubois, Li, Guestrin,
  Liang, and Hashimoto}]{alpaca}
Rohan Taori, Ishaan Gulrajani, Tianyi Zhang, Yann Dubois, Xuechen Li, Carlos
  Guestrin, Percy Liang, and Tatsunori~B. Hashimoto. 2023.
\newblock Stanford alpaca: An instruction-following llama model.
\newblock \url{https://github.com/tatsu-lab/stanford_alpaca}.

\bibitem[{Touvron et~al.(2023)Touvron, Lavril, Izacard, Martinet, Lachaux,
  Lacroix, Rozière, Goyal, Hambro, Azhar, Rodriguez, Joulin, Grave, and
  Lample}]{touvron2023llama}
Hugo Touvron, Thibaut Lavril, Gautier Izacard, Xavier Martinet, Marie-Anne
  Lachaux, Timothée Lacroix, Baptiste Rozière, Naman Goyal, Eric Hambro,
  Faisal Azhar, Aurelien Rodriguez, Armand Joulin, Edouard Grave, and Guillaume
  Lample. 2023.
\newblock \href {http://arxiv.org/abs/2302.13971} {Llama: Open and efficient
  foundation language models}.

\bibitem[{Zhang* et~al.(2020)Zhang*, Kishore*, Wu*, Weinberger, and
  Artzi}]{Zhang*2020BERTScore:}
Tianyi Zhang*, Varsha Kishore*, Felix Wu*, Kilian~Q. Weinberger, and Yoav
  Artzi. 2020.
\newblock \href {https://openreview.net/forum?id=SkeHuCVFDr} {Bertscore:
  Evaluating text generation with bert}.
\newblock In \emph{International Conference on Learning Representations}.

\end{thebibliography}
\bibliographystyle{acl_natbib}
\newpage
\newpage
\appendix
\section{Appendix}
\label{appendix}
\subsection{Prompts used for Question Generation}
\label{app:question_generation}
Following the analysis from prior works \cite{kocisky-etal-2018-narrativeqa,pang-etal-2022-quality}, we formulate our question generation method as a two-step process: (1) Summarization and (2) Question generation from summary. 
In the first step, we design prompt for generating summary as shown below: \\
\texttt{\small \textit{Summarize the paragraphs below in the context of \{title\} in \{domain\}.}} \\
In the next step, we ask ChatGPT to generate questions from summary as shown below: \\
\texttt{\small \textit{Using the context below, come up with follow-up questions. Rank the generated questions in the order of decreasing complexity to answer and display only the top 3. 
\{context\}}}

To demonstrate the usefulness of our question generation process, we also establish a baseline with the same instructions where questions are directly generated from the passage. The prompt used for the baseline is:\\

\texttt{\small \textit{Using the context below, come up with three questions. Rank the generated questions in the order of decreasing complexity to answer and display only the top 3. 
\{context\}}} \\

\subsection{Question Generation using open source LLMs}
\label{app:alpaca_QG}
In order to create a fair evaluation setup, we prompt Alpaca-7B,13B models to summarize and generate questions on 50 instances. We do not consider question generation from non-instruction tuned models (e.g: Llama). From our evaluation method on generated question as described in Section \ref{sec:results}, we find questions generated from Alpaca to be unanswerable (non-existent in the context) and contain gibberish content more than 80\% of the time.  The below table presents our evaluation of question generation from Alpaca:

\begin{table}[h!]
\footnotesize
\resizebox{0.5\textwidth}{!}{%
\begin{tabular}{lll}
\toprule
\textbf{Evaluation Metric} & \textbf{QG - Passage} & \textbf{QG - Summary}\\
\hline
Q1: Unambiguity  & 12.5\% & 8.3\% \\
\midrule
Q2. Context Length:  &  & \\
A sentence or less than a paragraph  & 98.8\% & 98.5\% \\
At least a third or most of the passage & 1.2\% & 1.5\% \\ 
\midrule
Q3: Multi-pass of the passage & 0\% & 0\% \\
\bottomrule
\end{tabular}
}
\caption{Prompts designed to evaluate the complexity of generated questions on Alpaca.} 
\label{tab:prompts_QG_eval}
\end{table}

\subsubsection{Evaluation of Question Generation using ChatGPT}
In order to verify the complexity of generated questions as outlined in Section ~\ref{sec:results}, we prompt ChatGPT with the following prompt:
\texttt{\small We would like to request your feedback on determining the complexity of generated questions by an AI assistant with respect to the context displayed above.\textbackslash n\textbackslash n For each of the question, rate the complexity of each of the generated questions for the dimensions: ambiguity, context and reasoning capabilities. \textbackslash n\textbackslash n \textbf{Q1: Is the question answerable from the given context and is unambiguous? A. Yes B. No \textbackslash n\textbackslash n Q2. How much of the passage is needed as context to answer the question? A. Only a sentence or two from the passage B. More than 2 sentences but lesser than a paragraph C. Atleast a third of the entire context given D. Most of the context given \textbackslash n\textbackslash n Q3: Does the question require multiple passes through the passage? A. Yes B. No}. Assume you do not have prior knowledge about the topic apart from the context given to you. Please output your choices in the form of a dictionary. (e.g: {'Q1': '<your answer choice for Q1>', 'Q2': '<your answer choice for Q2>', 'Q3': '<your answer choice for Q3>', 'Q4': '<your answer choice for Q4>'}). \textbackslash n\textbackslash n In the subsequent line, please provide a comprehensive explanation of your evaluation, avoiding any potential bias and ensuring that the order in which the responses were presented does not affect your judgment.} \\

Since LLMs are trained with different training data mixtures, we specifically ask ChatGPT to answer the questions based on the given context alone. 

\subsubsection{Prompt for Answer Generation}
In order generate response on the questions generated by LLMs, we prompt the following:
For every generated question, we prompt the models as follows: \\
\texttt{\small Given the context, answer the question below: \\
Context: \{context\} \\
Question: \{question\} \\
Answer: \{Answer\} \\
}

\subsection{GPT-4 as an Evaluator}
\subsubsection{Prompts used in GPT-4 Evaluation}
\label{app:answer_evaluation}
In order to evaluate answers generated by LLMs, we ask GPT-4 to rate answers on Likert scale from 0 to 3 (higher the better) on all of the following four metrics: coherency, relevance, accuracy and factual consistency. Our evaluation prompt used as prompt for GPT-4 is shown below:

\texttt{\small \textit{system prompt:} You are a helpful and precise assistant for checking the quality of the answer on 4 verticals: coherence, relevance, factual consistency, accuracy.
prompt : We would like to request your scores and feedback on the performance of two AI assistants for answering the user question based on the context displayed above. Please rate the answer quality on 4 metrics: coherence, relevance, factual consistency and accuracy. Definition of each metric is given to you. \textbf{Coherence} - Answer should be well-structured and well-organized. \textbf{Relevance} - Answer should be relevant to the question and the context. Answer should also avoid drifting from the question being asked. \textbf{Factual consistency} - The context should be the primary source for the answer. The answer should not contain fabricated facts and should entail information present in the context. \textbf{Accuracy} - Does the response provided by the assistant answer the question correctly in a concise manner? Provide a score to each AI assistant response for each of the metric defined above on a scale of 0 to 3 where higher score means better performance. Do not compare the performance between AI assistants and rate them individually. Enter two new lines. Next, please provide a comprehensive explanation of your preferred answer among the both and your evaluation, avoiding any potential bias and ensuring that the order in which the responses were presented does not affect your judgment. Please note that your scores should be in a dictionary format as the example given to you. Example: {'Assistant 1': {'coherence': <int score>, 'relevance': <int score>, 'factual consistency': <int score>, 'accuracy': <int score>}, 'Assistant 2': ...}  'Explanation': '<Preference: Assistant 1 or Assistant 2 or Both or Neither>.  <your explanation>'.}

\subsection{Case Study: Human Eval v.s GPT-4}
\label{app:human_eval}
We annotate 50 QA pairs to evaluate answer quality generated by LLMs independently of GPT-4 ratings. Two annotators are given similar guidelines as outlined in Section~\ref{sec:results} and a questionnaire as GPT-4 (Section~\ref{app:answer_evaluation}). For fair evaluation, we do not reveal the model that generated the answer to annotators. Table~\ref{tab:cohen_kappa} includes the agreement scores of the human evaluation with GPT-4. A high agreement score across different metrics on free-form text generation shows that our evaluation method using GPT-4 is reliable.
\begin{table}[]
    \centering
    \begin{tabular}{cc}
    \toprule
     Metric    & Cohen-Kappa score \\
     \midrule
      Coherency   & 0.66 \\
      Relevance & 0.7 \\
      Accuracy & 0.66 \\
      Factual Consistency & 0.56 \\
      \bottomrule
    \end{tabular}
    \caption{Annotator agreement scores with GPT-4}
    \label{tab:cohen_kappa}
\end{table}
We also present some qualitative examples in Appendix (Table~\ref{tab:examples_appendix})

\subsection{Data Collection}
\label{app:data_source}
We use Wikipedia API to collect articles to create our evaluation setting. However, we find not all articles may be useful to our setting; such as Wikipedia articles containing reference links, list of topics, glossary etc. We apply two levels of filtering: (1) We remove any article that has < 2k tokens (2) we curate a list of non-informative sections and filter those sections from consideration in this study.
\texttt{['References', 'See also', 'Notes', 'Sources', 'External links', 'History', 'Further reading']
    }
\subsection{Qualitative Examples}
Below table presents qualitative examples from human annotation task and shows some of the ratings by humans and GPT-4.
\begin{table*}[h!]
\footnotesize
\begin{tabular}{p{\textwidth}}
\hline
\\
\textbf{\textit{Qualitative Examples}} \\
\midrule
Question 1: What is the difference between interdisciplinary study and specialization in a single discipline? \\
\midrule
\textbf{Evaluation: Issue - None, Human score - 3, GPT-4 - 3} \\
\colorbox{yellow}{ChatGPT: } Interdisciplinary study involves applying the methods and approaches of several different disciplines to a subject, while specialization in a single discipline involves in-depth study and expertise in a single subject.\\
\\
\textbf{Evaluation: Issue - Repetition, Human score - 2, GPT-4 - 2} \\
\colorbox{yellow}{Alpaca-13B: } Interdisciplinary study is the combination of two or more disciplines to create a new field of study. Specialization is the focus on a single discipline, with the goal of becoming an expert in that field. \colorbox{lime}{Interdisciplinary study is the combination of two or more disciplines to create a new field of study.} \colorbox{lime}{Specialization is the focus on a single discipline, with the goal of becoming an expert in that field.} \\
\midrule
Question 2: How does good asset management require leadership, communication, collaboration, and continuous improvement? \\
\midrule
 \textbf{Evaluation: Issue - None, Human score - 3, GPT-4 - 3} \\
\colorbox{yellow}{Alpaca-13B: } Good asset management requires leadership, communication, collaboration, and continuous improvement.  Leadership is required to set the strategic priorities and to ensure that the organization is aligned with the goals. Communication is required to ensure that the organization is aware of the goals and the strategies to achieve them. Collaboration is required to ensure that the organization is working together to achieve the goals. Continuous improvement is required to ensure that the organization is constantly striving to improve its performance. \\
\\
\textbf{Evaluation: Issue - Accuracy, Human score - 1, GPT-4 - 1} \\ 
\colorbox{yellow}{Alpaca-7B: } Good asset management requires leadership, communication, collaboration, and continuous improvement. 
\\
\midrule
Question 3: What are the four areas that CSAB considers crucial to the discipline of computer science? \\
\midrule
\textbf{Evaluation: Issue - Factual Consistency, Human score - 1, \colorbox{pink}{GPT-4 - 3}} \\
\colorbox{yellow}{Alpaca-7B: }  1. Theory of computation \colorbox{pink}{2. Computability theory} 3. Programming language theory \colorbox{pink}{4. Formal methods} \\
\\
\colorbox{yellow}{ChatGPT: } The four areas that CSAB considers crucial to the discipline of computer science are theory of computation, algorithms and data structures, programming methodology and languages, and computer elements and architecture.
 \\
 
\hline
\end{tabular}
\caption{Few examples of generated answers from LLMs with GPT-4 and human evaluation
\\ \textit{Legend} \colorbox{yellow}{LLM}, \colorbox{pink}{Inconsistencies missed by the model} \colorbox{lime}{Inconsistencies detected by GPT-4.}}
\label{tab:examples_appendix}
\end{table*}
\end{document}